\documentclass[10pt,letterpaper]{article}
\usepackage[top=0.85in,left=2.75in,footskip=0.75in,marginparwidth=2in]{geometry}

\usepackage[utf8]{inputenc}

\usepackage{cite}

\usepackage{nameref,hyperref}

\usepackage{microtype}
\DisableLigatures[f]{encoding = *, family = * }

\raggedright
\usepackage{changepage}

\usepackage[aboveskip=1pt,labelfont=bf,labelsep=period,singlelinecheck=off]{caption}

\makeatletter
\renewcommand{\@biblabel}[1]{\quad#1.}
\makeatother

\usepackage{lastpage,fancyhdr,graphicx}
\usepackage{epstopdf}
\fancyheadoffset[L]{2.25in}
\fancyfootoffset[L]{2.25in}

\usepackage{color}

\definecolor{Gray}{gray}{.25}

\usepackage{graphicx}

\usepackage{sidecap}

\usepackage{amsmath}
\usepackage{amssymb}
\usepackage{xfrac}
\usepackage[pscoord]{eso-pic}
\usepackage[fulladjust]{marginnote}
\reversemarginpar

\begin{document}
\vspace*{0.35in}

\begin{flushleft}
{\Large
\textbf\newline{Neural networks with dynamical coefficients and adjustable connections on the basis of integrated backpropagation}
}
\newline
\\
M. N. Nazarov\textsuperscript{1,*},
\\
\bigskip
\bf{1} National Research University of Electronic Technology, Moscow.
\\
\bigskip
* nazarov-maximilian@yandex.ru

\end{flushleft}

\section*{Abstract}
We consider artificial neurons which will update their weight coefficients with an internal rule based on backpropagation, rather than using it as an external training procedure.
To achieve this we include the backpropagation error estimate as a separate entity in all the neuron models and perform its exchange along the synaptic connections.
In addition to this we add some special type of neurons with reference inputs, which will serve as a base source of error estimates for the whole network.
Finally, we introduce a training control signal for all the neurons, which can enable the correction of weights and the exchange of error estimates.
For recurrent neural networks we also demonstrate how to integrate   backpropagation through time into their formalism with the help of some stack memory for reference inputs and external data inputs of neurons.
Also, for widely used neural networks, such as  long short-term memory, radial basis function networks, multilayer perceptrons and convolutional neural networks, we demonstrate their alternative description within the framework of our new formalism.
As a useful consequence, our approach  enables us to introduce neural networks with the adjustment of synaptic connections, tied to the integrated backpropagation.

\section*{Introduction}

Backpropagation is one of the most successful and widely used algorithms for the training of neural networks. It has been adapted for such diverse models as multi-layer perceptrons, radial basis function networks and convolutional neural networks [\ref{Dreyfus}, \ref{Broomhead}, \ref{Lecun}]. 
Moreover, its modification of backpropagation through time (BPTT) has been successfully applied for the training of specialized recurrent neural networks, such as long short-term memory  [\ref{Greff}, \ref{Chen}]. 
The range of applied tasks that can be solved by these models is also quite diverse.
For example, convolutional networks are used [\ref{Krizhevsky}, \ref{Girshick}] for image recognition, networks of radial basis functions are used for time series prediction and control systems construction [\ref{Park}], networks of long short-term memory are used for a handwritten text recognition and generation [\ref{Pham}, \ref{Graves}], machine translation [\ref{Sutskever}], speech synthesis and recognition [\ref{Sak}, \ref{Fan}], and for a video processing in conjunction with convolutional networks [\ref{Donahue}].

However, the implementation of backpropagation is an external training procedure in relation to the models considered.
Therefore, if we want to build a network with dynamical coefficients (see examples in [\ref{NazarNets1}]) on the basis of this algorithm, we will need to include it directly into the core formalism of standard models of neurons.

This entails the introduction of backpropagation error estimates $\Delta(t)$ as some  separate entities, as well as  special neurons $N_{e}$ with reference inputs $e(t)$ and a training control signal $a(t)$ for our network. In the case of recurrent networks we will have to add stack memory $S_{x}$ for external data inputs and $S_{e}$ for reference inputs of neurons.

As a result, our networks could be viewed as a special type of reprogrammable finite automata. The first consequence is an ability to construct hierarchical networks, which will control the training process for one another in ascending order. As a simple example one can consider two networks: the first is trained to spot some special stimuli in the input data to activate the training of a much bigger second one and control which parts of data will be sent to its data inputs and which to its reference inputs.
Another important consequence would be the ability to introduce neural networks with the adjustment of synaptic connections (see the review in \cite{Maslennikov}) on the basis of integrated backpropagation. In theory the ideal connection adjustment algorithm should prevent the overfitting of data by deleting all the unused connections and creating new links only when necessary.

%
%

\section{The description of basic models}

A neuron number $j$ from layer number $i$ will be denoted as $N^{i\, j}_{\ldots}$.
Subscripts for $N^{i\, j}_{\ldots}$ will be variable-length strings: $\bf \varphi,r,c,e$, where $\varphi$ is an activation function, $\bf r $ denotes a recurrent mode, $\bf c$~identifies a mode with connection adjustment, and $\bf e$ denotes the presence of reference input for that neuron.
In the case of a non-recurrent neuron with static connections and without the reference input, only $\varphi$ will remain in this string.  For example, the notation $N^{i\, j}_{\sigma}$ will specify an ordinary neuron with a sigmoid activation function.
In the general case we will introduce neurons $N^{i\, j}_{\ldots}$ in our models as:
\[
	N^{i\, j}_{\ldots e}(t) = \left(  	\overline{c}^{i\, j}(t),
							\overline{x}^{i\, j}(t),
							\overline{\omega}^{i\, j}(t),
							b^{i\, j}(t),
							{\psi}^{i\, j},
							{\varphi}^{i\, j},
							\overline{y}^{i\, j}(t),
							{a}^{i\, j}(t),
							\overline{\Delta}^{i\, j}(t),
							{p}^{i\, j}(t),
							{\xi}^{i\, j}(t),
							\overline{e}^{i\, j}(t)
							 \right).
\]
\begin{itemize}
\item $\overline{c}^{i\, j}(t) = ({c}^{i\, j}_{1}(t),\ldots,{c}^{i\, j}_{n}(t))$ --- connections to other layers and external inputs.
\begin{enumerate}
	\item If the input $ k $ is not connected to anything, then  ${c}^{i\, j}_{k}(t) = (0,0,0)$.
	\item If the input $ k $ is connected to the external input  $X_{m}(t)$, then  ${c}^{i\, j}_{k}(t) = (0,0,m)$.
	\item If the input $ k $ is connected to the output $ r $ of a neuron $N^{l\, m}_{\ldots}$, then ${c}^{i\, j}_{k}(t) = (l,m,r)$.
\end{enumerate}
\item $\overline{x}^{i\, j}(t) = ({x}^{i\, j}_{1}(t),\ldots,{x}^{i\, j}_{n}(t))$ --- data input values of $N^{i\, j}_{\ldots e}(t)$.
\item $\overline{\omega}^{i\, j}(t) = ({\omega}^{i\, j}_{1}(t),\ldots,{\omega}^{i\, j}_{n}(t))$ ---  weight coefficients of $N^{i\, j}_{\ldots e}(t)$.
\item $b^{i\, j}(t)$ --- bias of neuron $N^{i\, j}_{\ldots e}(t)$.
\item ${\psi}^{i\, j}$ --- aggregation function of $N^{i\, j}_{\ldots e}(t)$, for example ${\psi}(\overline{\omega},\overline{x}) = \sum \omega_{k}\cdot x_{k}+b$.
\item ${\varphi}^{i\, j}$ --- activation function of $N^{i\, j}_{\ldots e}(t)$, for example $\varphi (z) = \mathrm{th} (z)$.
\item $\overline{y}^{i\, j}(t) = ({y}^{i\, j}_{1}(t),\ldots,{y}^{i\, j}_{k}(t))$ --- output values $\overline{y}^{i\, j}(t) = \varphi (\psi (\overline{\omega}^{i\, j}(t),\overline{x}^{i\, j}(t))) $.
\item $a^{i\, j}(t)$ --- input signal of training activation for $N^{i\, j}_{\ldots e}(t)$.
\item $\overline{\Delta}^{i\, j}(t) = ({\Delta}^{i\, j}_{1}(t),\ldots,{\Delta}^{i\, j}_{n}(t))$ --- coefficients for backpropagation from $N^{i\, j}_{\ldots e}(t)$.
\item $p^{i\, j}(t)$ --- paralysis indicator for weights of $N^{i\, j}_{\ldots e}(t)$.
\item $\xi^{i\, j}(t)$ --- local minimum indicator of $N^{i\, j}_{\ldots e}(t)$.
\item $\overline{e}^{i\, j}(t) = ({e}^{i\, j}_{1}(t),\ldots,{e}^{i\, j}_{k}(t))$ --- optional reference inputs for $N^{i\, j}_{\ldots e}(t)$.
\end{itemize}

\textbf{Remark 1}: For data inputs   of neurons four modes of operation are allowed: 
\begin{enumerate}
\item[1)] when ${c}^{i\, j}_{k}(t) = (0,0,0)$, we will have the zero input $x^{i\, j}_{k}(t) = 0$;
\item[2)] when ${c}^{i\, j}_{k}(t) = (0,0,m)$, we will have external connection $x^{i\, j}_{k}(t) = X_{m}(t)$;
\item[3)] for ${c}^{i\, j}_{k}(t) = (l,m,r)$ and $l< i$, we will have an ordinary link $x^{i\, j}_{k}(t) =y^{l\, m}_{r}(t) $;
\item[4)] for ${c}^{i\, j}_{k}(t) = (l,m,r)$ and $l \geq i$, we will have a recurrent one $x^{i\, j}_{k}(t) =y^{l\, m}_{r}(t-1) $.
\end{enumerate}

 \subsection{ Standard neurons and elementary computational blocks }

\textbf{\underline{Model 1}}: Neurons $N^{i\, j}_{\sigma}$ and $N^{i\, j}_{\sigma e}$ with sigmoid activation $ \varphi^{i\, j}(z) = \sfrac{1}{(1+e^{-2 \alpha z})}$.
The derivative of this function is: $\sfrac{\partial \varphi}{\partial z} = 2 \alpha \varphi(z) (1 - \varphi(z)) $. As an aggregation function we will use a weighted summation with the bias ${\psi^{i\, j}}(\overline{\omega},\overline{x}) = \sum \omega_{k}\cdot x_{k} + b$,
which as a result gives us a standard formula for  $y^{i\, j}(t) =\varphi^{i\, j} \left( \sum \omega^{i\, j}_{k}(t)\cdot x^{i\, j}_{k}(t) + b^{i\, j}(t) \right) $. Finally, general correction factors $\delta^{i\, j}(t)$   will be calculated:
 \begin{gather}\label{nazarov:neuro2:delta}
	\delta^{i\, j}(t) =  \left\lbrace  \begin{aligned} (y^{i\, j}(t) - e^{i\, j}(t)), \quad & \text{for  neurons with}\,\, e^{i\,j}; \text{ denote them by } N^{i\, j}_{\sigma e}; \\
	\sum_{ \substack{l,p,k:\\ c^{l\, p}_{k}(t)=(i,j,1)  }} \Delta^{l\, p}_{k}(t), \quad & \text{for  neurons without}\,\, e^{i\,j}; \text{ denote them by } N^{i\, j}_{\sigma}. \end{aligned} \right.
 \end{gather}
The application of formula  \ref{nazarov:neuro2:delta} implies the explicit inclusion of weights $\omega_{k}^{i\, j}(t)$ into all of the backpropagation coeffitients $\Delta_{k}^{i\, j}(t)$, which  yields:
\begin{gather}\label{nazarov:neuro2:Delta}
 \Delta_{k}^{i\, j}(t) = 2 \alpha \, y^{i\, j}(t) \, \left(1 - y^{i\, j}(t) \right) \, \delta^{i\, j}(t) \, \omega_{k}^{i\, j}(t) \, \sigma(a^{i\, j}(t)).
\end{gather}
The activation of training will be applied with a positive training signal  $a^{i\, j}(t) > 0$. For the adjustment of weights we will use a standard formula with an added $ \sigma(a^{i\, j}(t))$:
\begin{gather}\label{nazarov:neuro2:omega}
 \omega_{k}^{i\, j}(t+1) = \omega_{k}^{i\, j}(t) - 2 \mu\, \alpha\, y^{i\, j}(t)  \, \left(1 - y^{i\, j}(t) \right)\, \delta^{i\, j}(t)\,   x_{k}^{i\, j}(t)  \, \sigma(a^{i\, j}(t)).
\end{gather}
Assuming $x_{k}^{i\, j}(t)=1$ we will get a formula for bias  adjustment $b^{i\, j}(t+1)$ from \ref{nazarov:neuro2:omega}.\\

We assume that a paralysis of weights $\omega_{k}^{i\, j}(t)$ occurs when 70\,\% of them reach a threshold  value $\omega_{\max}$:
\begin{gather}\label{nazarov:neuro2:paralysis}
p^{i\, j}(t) = \left\lbrace \begin{aligned}
				1, & \quad \text{if} \, \sum_{k = \overline{ 1,n}} |\omega_{k}^{i\, j}(t)| > 0.7\cdot \omega_{\max} \cdot n, \\
				0, & \quad \text{if otherwise}.
				\end{aligned}  \right.
\end{gather}
In this expression, $ n$ is the number of neuron inputs $x^{i\, j}_{1},\ldots, x^{i\, j}_{n}$. Formulas for detecting a local minimum of $\omega_{k}^{i\, j}(t)$ will also use this number, but the main criteria for them will be a low amplitude oscillation of $\Delta \omega_{k}^{i\, j}(t) =\omega_{k}^{i\, j}(t+1) - \omega_{k}^{i\, j}(t) $:
\begin{gather}\label{nazarov:neuro2:localmin}
\xi^{i\, j}(t) = \left\lbrace \begin{aligned}
				1, & \quad \text{if} \,   \sum_{k = \overline{ 1,n}} \left| \sum_{\tau = \overline{t-t_{\xi},t}}   \omega_{k}^{i\, j}(\tau+1) - \omega_{k}^{i\, j}(\tau) \right| < \omega_{\min} \cdot  n\cdot \! \! \! \prod_{\tau = \overline{t-t_{\xi},t}} \sigma (a^{i\, j}(\tau)), \\
				0, & \quad \text{if otherwise}.
				\end{aligned}  \right.
\end{gather}
As a result, our basic model of sigmoid neuron  will have only six parameters:
\begin{itemize}
\item $n$ --- number of data inputs;
\item $\omega_{\max}$ --- maximum absolute values of weights $\omega^{i\, j}_{k}$;
\item $\omega_{\min}$ --- minimum absolute values of weights $\omega^{i\, j}_{k}$;
\item $t_{\xi}$ --- local minimum detection time;
\item $\mu$ --- training rate of neuron;
\item $\alpha$ --- sigmoid stiffness ($\alpha \geq 1$).
\end{itemize}

 
\noindent \textbf{\underline{Model 2}}: Neurons $N^{i\, j}_{\mathrm{th}}$ and $N^{i\, j}_{\mathrm{th}\, e}$ with hyperbolic tangent as an activation function~$ \varphi^{i\, j}(z) = \mathrm{th}(z) $. We will use a weighted summation with the bias ${\psi^{i\, j}}(\overline{\omega},\overline{x}) = \sum \omega_{k}\cdot x_{k} +b$ as an aggregation function $\psi^{i\, j}$, just as in model 1.
The derivative of the activation function will be $\sfrac{\partial \varphi}{\partial z} =  (1 - \varphi^{2}(z)) $ which leads to the replacement of formulas   \ref{nazarov:neuro2:Delta} and \ref{nazarov:neuro2:omega}:
\begin{gather}\label{nazarov:neuro2:Deltath}
 \Delta_{k}^{i\, j}(t) = \left(1 - (y^{i\, j}(t))^{2} \right) \, \delta^{i\, j}(t) \, \omega_{k}^{i\, j}(t) \, \sigma(a^{i\, j}(t)),
\end{gather}
\begin{gather}\label{nazarov:neuro2:omegath}
 \omega_{k}^{i\, j}(t+1) = \omega_{k}^{i\, j}(t) -   \mu\, \left(1 - (y^{i\, j}(t))^{2} \right)\, \delta^{i\, j}(t)\,   x_{k}^{i\, j}(t)  \, \sigma(a^{i\, j}(t)).
\end{gather}
Bias update $b^{i\, j}(t+1)$ is a special case of \eqref{nazarov:neuro2:omegath} and can be obtained by a simple substitution $x_{k}^{i\, j}(t) = 1$. Moreover, the formulas for calculating $\delta^{i\, j}(t)$,  $p^{i\, j}(t)$, and $\xi^{i\,j}(t)$ are completely analogous to   \ref{nazarov:neuro2:delta},  \ref{nazarov:neuro2:paralysis} and \ref{nazarov:neuro2:localmin}.

\noindent \textbf{\underline{Model 3}}: Neurons $N^{i\, j}_{\mathrm{id}}$ and $N^{i\, j}_{\mathrm{id}\, e}$ with a linear  activation function $ \varphi^{i\, j}(z) = z $. Just as in the first two models, we will use the standard aggregation function ${\psi^{i\, j}}(\overline{\omega},\overline{x}) = \sum \omega_{k}\cdot x_{k} +b$.
Formulas for $\delta^{i\, j}(t)$, $p^{i\, j}(t)$, and $\xi^{i\,j}(t)$ will be analogous to \ref{nazarov:neuro2:delta}  \ref{nazarov:neuro2:paralysis} and \ref{nazarov:neuro2:localmin}.  
In turn, an expression for $\Delta^{i\,j}_{k}(t)$ and $\omega^{i\,j}_{k}(t+1)$ considering the linear $\varphi^{i\, j}$ will be replaced by
 \begin{gather*}\label{nazarov:neuro2:Deltaid}
 \Delta_{k}^{i\, j}(t) =  \delta^{i\, j}(t) \, \omega_{k}^{i\, j}(t) \, \sigma(a^{i\, j}(t)),
\end{gather*}
\begin{gather*}\label{nazarov:neuro2:omegaid}
 \omega_{k}^{i\, j}(t+1) = \omega_{k}^{i\, j}(t) -   \mu\,  \delta^{i\, j}(t)\,   x_{k}^{i\, j}(t)  \, \sigma(a^{i\, j}(t)).
\end{gather*}


\noindent\textbf{\underline{Model 4}}: Neurons $N^{i\, j}_{\mathrm{Ed}}$ and $N^{i\, j}_{\mathrm{Ed}\, e}$  for the calculation of Euclidean distance, which use $\varphi^{i\, j}(z) = \sqrt{z}$ as an activation function and ${\psi^{i\, j}}(\overline{\omega},\overline{x}) = \sum (\omega_{k} - x_{k})^{2}$ as an aggregation function. As a result, an output value for them is  $y^{i\,j}(t) = \sqrt{ \sum \left(\omega^{i\,j}_{k}(t) - x^{i\,j}_{k}(t)\right)^{2}}$.
General correction factors $\delta^{i\, j}(t)$ will be:
 \begin{gather*}\label{nazarov:neuro2:deltaEd}
	\delta^{i\, j}(t) =\left\lbrace  \begin{aligned} \dfrac{1}{2}(y^{i\, j}(t) - e^{i\, j}(t)), \quad & \text{for neurons with}\,\, e^{i\,j}; \text{ denote them by } N^{i\, j}_{\mathrm{Ed}\, e}; \\
	\sum_{ \substack{l,p,k:\\ c^{l\, p}_{k}(t)=(i,j,1)  }} \Delta^{l\, p}_{k}(t), \quad & \text{for neurons without}\,\, e^{i\,j}; \text{ denote them by } N^{i\, j}_{\mathrm{Ed}}. \end{aligned} \right.
 \end{gather*}
Taking into account a special aggregation function, we will have the following formula for backpropagation coefficients $\Delta_{k}^{i\,j}(t)$:
\begin{gather*}\label{nazarov:neuro2:DeltaEd}
 \Delta_{k}^{i\, j}(t) = \delta^{i\, j}(t) \cdot \left( \omega_{k}^{i\, j}(t) - x_{k}^{i\, j}(t) \right) \cdot \sigma(a^{i\, j}(t)) / y^{i\, j}(t).
\end{gather*}
Finally, a formula for weight coefficients $ \omega_{k}^{i\, j}(t+1)$ will be updated as follow:
\begin{gather*}\label{nazarov:neuro2:omegaEd}
 \omega_{k}^{i\, j}(t+1) =    \omega_{k}^{i\, j}(t) - 2\mu\cdot \Delta_{k}^{i\, j}(t).
\end{gather*}
  

\noindent\textbf{\underline{Model 5}}: Convolutional neurons  $N^{i\, j}_{\mathrm{Conv}}$ and $N^{i\, j}_{\mathrm{Conv}\, e}$
with linear activation $ \varphi^{i\, j}(z) = z $, matrix input
 $\overline{x}^{i\, j}(t) = \left( {x}^{i\, j}_{1 1}(t),\ldots , {x}^{i\, j}_{n m}(t) \right)$, vector output $\overline{y}^{i\, j}(t) = \left({y}^{i\, j}_{1}(t),\ldots ,{y}^{i\, j}_{m}(t)  \right)$ and weighted summation as an aggregation ${\psi^{i\, j}}(\overline{\omega},\overline{x}) = \left( \sum \omega_{k}\cdot x_{k 1},\ldots ,\sum\omega_{k}\cdot x_{k m} \right)  $.
This variant will yield as its output the dot product of input data with the kernel of the weight coefficients
${\overline{\omega}^{i\,j}(t) =( \omega^{i\,j}_{1}(t),\ldots ,\omega^{i\,j}_{n}(t))}$.
Thus, the final output  values would be
$$\overline{y}^{i\, j}(t) = \left({y}^{i\, j}_{1}(t),\ldots ,{y}^{i\, j}_{m}(t)  \right) = \left( \sum\limits_{k=\overline{1,n}} \omega^{i\, j}_{k}(t) x^{i\, j}_{k 1}(t),\ldots ,\sum\limits_{k=\overline{1,n}}\omega^{i\, j}_{k}(t) x^{i\, j}_{k m}(t) \right).$$
As a result, general correction factors $\delta^{i\, j}(t)$ will be:
 \begin{gather*}\label{nazarov:neuro2:delta5}
	\delta^{i\, j}(t) = \left\lbrace  \begin{aligned}\sum_{r} (y_{r}^{i\, j}(t) - e_{r}^{i\, j}(t)), \quad & \text{for neurons with}\,\, \overline{e}^{i\,j}; \text{ denote them by } N^{i\, j}_{\mathrm{Conv}\, e}; \\
\sum_{r}	\!\!\!\!\! 	\sum_{ \substack{l,p,k:  \\ c^{l\, p}_{k}(t)=(i,j,r)  }} \Delta^{l\, p}_{k}(t) \quad & \text{for neurons without}\,\, \overline{e}^{i\,j}; \text{ denote them by }  N^{i\, j}_{\mathrm{Conv}}. \end{aligned} \right.
 \end{gather*}
We have used a double range of indices for the inputs $\overline{x}^{i\, j}(t) = \left( {x}^{i\, j}_{1 1}(t),\ldots , {x}^{i\, j}_{n m}(t) \right)$, which made it quite convenient to describe the aggregation operation. However, the application of a similar scheme for $\overline{c}^{i\, j}(t)$ and $\overline{\Delta}^{i\, j}(t)$ would break the compatibility with other neuron layers, which use the notation of models 1--4.
As a result, we will represent them as vectors $\overline{c}^{i\, j}(t) = ({c}_{1}^{i\, j}(t),\ldots ,{c}_{n\cdot m}^{i\, j}(t))$ and $\overline{\Delta}^{i\, j}(t) = ({\Delta}_{1}^{i\, j}(t),\ldots ,{\Delta}_{n\cdot m}^{i\, j}(t))$, while   binding them with inputs $\overline{x}^{i\, j}(t)$  ($k_{1} = \overline{1,n}$ and $k_{2} = \overline{1,m}$):
\begin{itemize}
\item when in the disconnected mode $c_{(k_{1}-1)m+k_{2}}^{i\,j} = (0,0,0) \longrightarrow x^{i\,j}_{k_{1} k_{2}} = 0$;
\item for the external input $c_{(k_{1}-1)m+k_{2}}^{i\,j} = (0,0,r) \longrightarrow x^{i\,j}_{k_{1} k_{2}}(t) = X_{r}(t)$;
\item in the standard mode $c_{(k_{1}-1)m+k_{2}}^{i\,j} = (l,p,r) \longrightarrow x^{i\,j}_{k_{1} k_{2}}(t) = y^{l\, p }_{r}(t) $ ($l<i$);
\item in the recurrent mode $c_{(k_{1}-1)m+k_{2}}^{i\,j} = (l,p,r) \longrightarrow x^{i\,j}_{k_{1} k_{2}}(t) = y^{l\, p }_{r}(t-1) $ ($l \geq i$).
\end{itemize}
For backpropagation coefficients, we will have the corresponding formula:
\begin{gather*}\label{nazarov:neuro2:Delta3}
 \Delta_{(k_{1}-1)m+k_{2}}^{i\, j}(t) = \delta^{i\, j}(t) \cdot   \omega_{k_{1}}^{i\, j}(t)   \cdot \sigma(a^{i\, j}(t)), \quad \text{where} \,\, k_{1} = \overline{1,n}\,\, \text{and} \,\, k_{2} = \overline{1,m}.
\end{gather*}
We will not be able to use general correction factors $\delta^{i\, j}(t) $ in the pure form for   weights coefficients, as a result our formula for their update would be quite complicated:
  \begin{gather}\label{nazarov:neuro2:omega3}
	\omega_{k}^{i\, j}(t+1) = \omega_{k}^{i\, j}(t)  -\mu \sum_{r} x^{i\,j}_{k r}   \left\lbrace  \begin{aligned}  (y_{r}^{i\, j}(t) - e_{r}^{i\, j}(t)),    \quad & \text{for} \,\, N^{i\, j}_{\mathrm{Conv}\, e}; \\
	  \! \sum_{ \substack{l_{1},l_{2},p: \\ c^{l_{1}\, l_{2}}_{p}(t)=(i,j,r)  }} \Delta^{l_{1}\, l_{2}}_{p}(t),  \quad & \text{for} \,\, N^{i\, j}_{\mathrm{Conv}}. \end{aligned} \right.
 \end{gather}
All other formulas for $p^{i\, j}(t)$ and  $\xi^{i\,j}(t)$ would be the same as in model 1.


\noindent\textbf{\underline{Model 6}}: Linear rectification blocks
$B^{i\, j}_{\mathrm{ReLu}}$ and $B^{i\, j}_{\mathrm{ReLu}\, e}$ with an activation function $\varphi^{i\,j}(x^{i\, j}(t)) =$ ${= \max(0,x^{i\, j}(t))}$.
We do not call this blocks  of artificial neurons, because they do not have weight coefficients $\omega^{i\, j}(t)$ and an aggregation function.
As a smooth approximation of a $\max(0,z)$ one can take $\varphi^{i\,j}(z) \approx (1/2\alpha) \log (1 + e^{2\alpha z})$, which has the following derivative: $\dfrac{\partial \varphi}{\partial z} \approx \dfrac{1}{1+e^{-2\alpha z}}$.
General correction coefficients $\delta^{i\, j}(t)$    are calculated by a formula similar to  \eqref{nazarov:neuro2:delta},
while the backpropagation coefficient  $\Delta_{1}^{i\,j}(t)$ is
\begin{gather*}\label{nazarov:neuro2:Delta4}
 \Delta_{1}^{i\, j}(t) =  \delta^{i\, j}(t) \cdot \sigma(a^{i\, j}(t))\, / \left(  1 + e^{-2\alpha x^{i\, j}(t)}\right).
\end{gather*} 


\noindent\textbf{\underline{Model 7}}: Pooling layers  $B^{i\, j}_{\mathrm{Pool}}$ and $B^{i\, j}_{\mathrm{Pool}\, e}$ with a linear activation function $\varphi(z) =z$ and subsampling as an aggregation function $\psi(\overline{x^{i\, j}(t)})=\max(x^{i\, j}_{1}(t),\ldots , x^{i\, j}_{n}(t))$. General correction coefficients will be calculated similarly to formula \ref{nazarov:neuro2:delta}, while for the backpropagation coefficients  $\Delta_{k}^{i\,j}(t)$ we will have
\begin{gather*}\label{nazarov:neuro2:Delta5}
 \Delta_{k}^{i\, j}(t) = \left\lbrace \begin{aligned}  \delta^{i\, j}(t) \cdot \sigma(a^{i\, j}(t)), & \quad \text{if} \,\, \psi(\overline{x^{i\, j}(t)})=  x^{i\, j}_{k}(t),\\ 0, &\quad \text{if otherwise}. \end{aligned} \right.
\end{gather*}


\noindent\textbf{\underline{Model 8}}: Gaussian blocks  $B^{i\, j}_{\mathrm{norm}}$ and $B^{i\, j}_{\mathrm{norm}\, e}$ with the normal activation function $\varphi^{i\,j}(z) = e^{-\beta z^{2}}$ and only one single data input. The derivative of this function is $\dfrac{\partial \varphi}{\partial z} = \varphi(z) (-2\beta \sqrt{\log(\beta)-\log(\varphi(z))})$. As a result, for a single backpropagation coefficient  $\Delta_{1}^{i\,j}(t)$ we will have the following formula:
   \begin{gather*}\label{nazarov:neuro2:Deltanorm}
	\Delta_{1}^{i\, j}(t) = -2\beta y^{i\, j}(t)  \sqrt{\log(\beta)-\log(y^{i\, j}(t))}\, \delta^{i\, j}(t) \, \sigma(a^{i\, j}(t)).
 \end{gather*}
General correction coefficients  $\delta^{i\, j}(t)$ will use the same formula as in \ref{nazarov:neuro2:delta}.
 

\noindent\textbf{\underline{Model 9}}: Multiplication blocks  $B^{i\, j}_{*}$ and $B^{i\, j}_{*\, e}$ with two data inputs $x_{1}^{i\,j}$ and $ x_{2}^{i\,j}$, a linear activation $\varphi(z) =z$ and an aggregation function $\psi(x_{1},x_{2}) = x_{1} \cdot x_{2}$. General correction coefficients  $\delta^{i\, j}(t)$ will use formula analogous to  \ref{nazarov:neuro2:delta} and the back\-propagation coefficients:
 \begin{gather*}\label{nazarov:neuro2:Deltamult}
 \Delta_{1}^{i\, j}(t) =  \delta^{i\, j}(t) \cdot x^{i\,j}_{2}(t) \cdot \sigma(a^{i\, j}(t)), \quad
 \Delta_{2}^{i\, j}(t) =  \delta^{i\, j}(t) \cdot x^{i\,j}_{1}(t) \cdot \sigma(a^{i\, j}(t)).
\end{gather*}
 

\noindent\textbf{\underline{Model 10}}: Summation blocks  $B^{i\, j}_{+}$ and $B^{i\, j}_{+\, e}$ with two data inputs $x_{1}^{i\,j}$ and $ x_{2}^{i\,j}$, a linear activation $\varphi(z) =z$ and an aggregation   $\psi(x_{1},x_{2}) = x_{1} + x_{2}$. General correction coefficients  $\delta^{i\, j}(t)$ will use a formula analogous to  \ref{nazarov:neuro2:delta} and the back\-propagation coefficients
\begin{gather*}\label{nazarov:neuro2:Deltasum}
	\Delta_{k}^{i\, j}(t) = \sigma(a^{i\, j}(t))  \delta^{i\, j}(t).
 \end{gather*}


\noindent\textbf{\underline{Model 11}}: Tangent activator blocks  $B^{i\, j}_{\mathrm{th}}$ and $B^{i\, j}_{\mathrm{th}\, e}$ with a single input and hyperbolic tangent as an activation function $\varphi(z) = \tanh( z)$.
General correction coefficients  $\delta^{i\, j}(t)$ will use a formula analogous to  \ref{nazarov:neuro2:delta} and the back\-propagation coefficients
    \begin{gather*}\label{nazarov:neuro2:DeltaBlockth}
	\Delta_{k}^{i\, j}(t) = \left(1 - (y^{i\, j}(t))^{2} \right)\, \sigma(a^{i\, j}(t))\,   \delta^{i\, j}(t).
 \end{gather*}


 \subsection{Recurrent neurons with an integrated stack memory}

For recurrent neurons without reference input $\overline{e}^{i\, j}(t)$, we will use the following scheme:
\[
	N^{i\, j}_{\ldots r}(t) = \left(  	\overline{c}^{i\, j}(t),
							\overline{x}^{i\, j}(t),
							\overline{S_{x}}^{i\, j}(t),							
							\overline{\omega}^{i\, j}(t),
							b^{i\, j}(t),
							{\psi}^{i\, j},
							{\varphi}^{i\, j},
							\overline{y}^{i\, j}(t),
							{a}^{i\, j}(t),
							\overline{\Delta}^{i\, j}(t),
							{p}^{i\, j}(t),
							{\xi}^{i\, j}(t)
							 \right).
\]
In this expression we introduce a stack memory $\overline{S_{x}}^{i\, j}(t)$ for those data inputs of neurons, which are connected to an external input source:  $c^{i\, j}_{k}(t) =(0,0,r)$,  $x^{i\, j}_{k}(t) = X_{r}(t)$.
The stack memory $\overline{S_{x}}^{i\, j}(t)$  will be a function according to a standard algorithm.
\begin{enumerate}
\item When $\sigma(a^{i\, j}(t)) = 0$, we make writing to $S_{x_{k}}^{i\, j}$ for all  $k$,  if $c^{i\, j}_{k}(t) =(0,0,r)$ and $r\neq 0$:
\begin{gather}\label{nazarov:neuro2:stackmem}
	\forall m =\overline{1,MaxM} \quad S_{x_{k}}^{i\, j}(m,t+1) = S_{x_{k}}^{i\, j}(m-1,t), \quad S_{x_{k}}^{i\, j}(0,t+1) = x^{i\,j}_{k}(t).
 \end{gather}
 \item When $\sigma(a^{i\, j}(t)) = 1$, we make reading from $S_{x_{k}}^{i\, j}$ for all $k$,  if $c^{i\, j}_{k}(t) =(0,0,r)$ and $r\neq 0$:
\begin{gather}\label{nazarov:neuro2:stackmem2}
	\forall m =\overline{1,MaxM} \quad S_{x_{k}}^{i\, j}(m-1,t+1) = S_{x_{k}}^{i\, j}(m,t), \quad S_{x_{k}}^{i\, j}(MaxM,t+1) = 0.
 \end{gather}
\end{enumerate}
After the inclusion of the stack memory, two formulas from standard neuron models should be updated with the highest priority: a formula for  $y^{i\, j}(t)$ and $\delta^{i\, j}(t)$. We will consider their change with the example of the model 2 with hyperbolic tangent.
 \begin{gather}\label{nazarov:neuro2:deltarecurrent}
	\delta^{i\, j}(t) = \left\lbrace
	\sum_{ \substack{l,p,k:\,\, l>i \\ c^{l\, p}_{k}(t)=(i,j,1)  }} \Delta^{l\, p}_{k}(t)  + \sum_{ \substack{l,p,k:\,\, l \leq i \\ c^{l\, p}_{k}(t)=(i,j,1)  }} \Delta^{l\, p}_{k}(t-1)  \right\rbrace.
 \end{gather}
In general, we can guarantee that such scheme will implement the standard algorithm of   backpropagation through time  if we replace a formula for $y^{i\, j}(t)$ by this one:
 \begin{gather}\label{nazarov:neuro2:sigmay}
	 y^{i\, j}(t) = \left\lbrace  \begin{aligned} \varphi^{i\, j} \left( \sum_{k} \omega_{k}^{i\, j}(t) x_{k}^{i\, j}(t) \right)\!, \quad & \text{when} \, \, \sigma(a^{i\, j}(t))=0,\\
	 \varphi^{i\, j}\left( \sum_{{ k:\, c^{i\, j}_{k}\neq (0,0,r)}} \!\!\!\!\!\! \omega_{k}^{i\, j}(t) x_{k}^{i\, j}(t) +\!\!\!\! \sum_{{ k:\, c^{i\, j}_{k}= (0,0,r)}}\!\!\!\!\!\! \omega_{k}^{i\, j}(t) S_{x_{k}}^{i\, j}(0,t) \right)\!, \quad & \text{when} \, \, \sigma(a^{i\, j}(t))=1.
	  \end{aligned} \right.
 \end{gather}
Taking into account the use of the stack memory, a transfer of the value $\Delta^{l\, p}_{k}(t-1)$ with a unit delay with $a^{i\, j}(t) =a^{i\, j}(t-1) > 0$ will be exactly the transfer of training data from the future (one stack of higher level), rather then from the past.
Completely by analogy, we will change a formula for  the weight coefficients $\omega^{i\, j}_{k}(t+1)$ to update:
 \begin{gather}\label{nazarov:neuro2:omegarecurrent}
 \omega_{k}^{i\, j}(t+1) = \omega_{k}^{i\, j}(t)\! -\! \mu \left(1 - (y^{i\, j}(t))^{2} \right) \delta^{i\, j}(t)  \sigma(a^{i\, j}(t)) \left\lbrace \begin{aligned}  x_{k}^{i\, j}(t), & \quad \text{if} \,\, c_{k}^{i\, j} \neq (0,0,r);  \\
  S_{x_{k}}^{i\, j}(0,t), & \quad  \text{if otherwise}. \end{aligned} \right.
\end{gather}
At the same time, a formula for the backpropagation coefficients $\Delta^{i\,j}_{k}(t)$ will be \ref{nazarov:neuro2:Deltath}, the same as in standard model, as well as a formula for bias update $b^{i\, j}(t+1)$.
For the neurons $N^{i\,j}_{\mathrm{th}\, r\, e}$ with reference inputs ${e}^{i\, j}$, we will also add the stack memory $S_{e}^{i\, j}$.
\begin{enumerate}
\item When $\sigma(a^{i\, j}(t)) = 0$, we make writing to $S_{e}^{i\, j}$:
\begin{gather}\label{nazarov:neuro2:stackmem-e}
	\forall m =\overline{1,MaxM} \quad S_{e}^{i\, j}(m,t+1) = S_{e}^{i\, j}(m-1,t), \quad S_{e}^{i\, j}(0,t+1) = e^{i\,j}(t).
 \end{gather}
 \item When $\sigma(a^{i\, j}(t)) = 1$, we make reading from $S_{e}^{i\, j}$:
\begin{gather}\label{nazarov:neuro2:stackmem-e2}
	\forall m =\overline{1,MaxM} \quad S_{e}^{i\, j}(m-1,t+1) = S_{e}^{i\, j}(m,t), \quad S_{e}^{i\, j}(MaxM,t+1) = 0.
 \end{gather}
\end{enumerate}
Formulas for correction of coefficients $\delta^{i\,j}(t)$ of neurons $N^{i\,j}_{\mathrm{th}\, r\, e}$ will be replaced by the following ones:
\begin{gather}\label{nazarov:neuro2:stackmem-deltae}
	\delta^{i\,j}(t) =    \left\lbrace (y^{i\, j}(t) - S^{i\,j}_{e}(0,t)) +  \sum_{ \substack{l,p,k:\,\, l \leq i \\ c^{l\, p}_{k}(t)=(i,j,1)  }} \Delta^{l\, p}_{k}(t-1) \right\rbrace.
\end{gather}
If a neuron $N^{i\,j}_{\mathrm{th}\, r\, e}$ is allowed to have connections to external data inputs $X_{r}(t)$, then it will use formulas \ref{nazarov:neuro2:sigmay} and \ref{nazarov:neuro2:omegarecurrent}, while otherwise  standard formulas from model 2 are used.

As a result, all recurrent neurons will have only one additional parameter:
\begin{itemize}
\item $MaxM$ --- depth of stack memory.
\end{itemize}
 Without any fundamental differences a recurrent mode could be introduced to all the other standard neuron models with inclusion of the stack memory.
 To prepare our recurrent network for training on $m\leq MaxM$ etalon values, one will have to supply this data values $\overline{X}(t),\ldots , \overline{X}(t+m)$ with corresponding reference values ${e}(t),\ldots , {e}(t+m)$, while holding the $a^{i\,j}(t) = \ldots = a^{i\,j}(t+m) = 0$. To complete one full cycle of training, we will need to turn the training signal high and hold it for an additional $m$ time steps $a^{i\, j}(t+m+1) = \ldots = a^{i\, j}(t+2m)= 1$.


 \subsection{Neurons with adjustable connections}

The general idea of our neural link adjustment is to remove those connections that are almost out of use and create new connections with the most active neurons of the previous layers, provided the training with current connections can lead to a paralysis of weight coefficients or to a fluctuation of their values near the local minimum.

\textbf{Algorithm of a new link creation.} On each iteration $t$ of the neuron $N^{i\,j}_{\ldots c}$ we make the following steps:\\
\underline{Step 1.} Check if a neuron is unable to fix $\delta^{i\,j}(t)$ with current $x_{k}^{i\,j}(t)$, even if it raises all the weight coefficients almost to $\omega_{\max}$ (we choose $0.7\, \omega_{\max}$ as a control value):
\begin{gather*}\label{nazarov:neuro2:connection_new}
	C^{i\,j}_{new_{1}}(t) = 1, \,\, \text{if}\, \left| y(t) - \varphi\left( b^{i\, j}(t) - \mathrm{sign} (\delta^{i\, j}(t)) \sum_{k} x^{i\, j}_{k}(t)  \, 0.7 \, \omega_{\max}    \right) \right| < \left| \delta^{i\, j}(t) \right|.
\end{gather*}
If this event is detected ($C^{i\,j}_{new_{1}}(t) = 1$) and the training signal is active ($a^{i\,j}(t) =1$), then we will go to step 2. Otherwise, we need to verify additionally that a sign of $\delta^{i\, j}(t)$ and $\delta^{i\, j}(t-1)$ is different, while  $x^{i\, j}_{k}(t)$ and $x^{i\, j}_{k}(t-1)$ are almost identical:
\begin{gather}\label{nazarov:neuro2:connection_new2}
	C^{i\,j}_{new_{2}}(t) = 1, \quad \text{if}\,\, \delta^{i\, j}(t\! -\! 1) \,   \delta^{i\, j}(t)  < 0 \,  \text{and}\, \sum_{k} | x^{i\, j}_{k}(t) \! -\! x^{i\, j}_{k}(t\! -\! 1)| < 0.1\, n\, x_{\max}.
\end{gather}
If this event is detected ($C^{i\,j}_{new_{2}}(t) = 1$) and the training signal is active ($a^{i\,j}(t) =1$), then we will go to step 2. Otherwise, we restart the algorithm and wait for a next iteration.\\
\underline{Step 2.} Among all the connections of our neuron  $N^{i\,j}_{\ldots c}$, we are looking for an empty one: ${c^{i\,j}_{k}(t) =}$ ${= (0,0,0)}$. If we managed to find some suitable $k$, then we will proceed to step 3. Otherwise, we restart the algorithm and wait for a next iteration.\\
\underline{Step 3.} With a probability $P_{deep 1}$ we will go to step 4. If a transition to step 4 was not carried out, then we will search for a suitable candidate $y^{i-1\,p}_{r}$ for a new connection from previous $i-1$ layer of a neural net. We will carry out this selection according to the following conditions:
\begin{itemize}
\item there is no current connection to $y^{i-1\,p}_{r}$ from $N^{i\,j}_{\ldots c}$: $\not{\exists}\, \acute{k}: c_{\acute{k}}^{i\,j}(t) = (i-1,p,r)  $;
\item among all the admissible candidates, we select the maximum modulo: $\max |y^{i-1\,p}_{r}(t)|$;
\item if more than one $y^{i-1\,p}_{r}$ was found, then we will choose any random one of them.
\end{itemize}
If we managed to find some suitable $ p $ and $ r $, then we will go directly to the final step 5; otherwise, we will go to step 4 first. \\
\underline{Step 4.} If our neuron is $N^{i\,j}_{\ldots r\, c}$ from a recurrent layer, then with a probability $P_{rec}$ we will try to create a recurrent connection; otherwise, we will try to create a deep connection with some distant neural layers.
Among all the  $y^{l\,p}_{r}$ ($l< i-1$ for a direct one and $l\geq i$ for a recurrent one), we choose such one that the following conditions hold:
\begin{itemize}
\item there is no current connection from $N^{i\,j}_{\ldots c}$: $\not{\exists}\, \acute{k}: c_{\acute{k}}^{i\,j}(t) = (l,p,r)  $ and $(l,p)\neq (i,j)$;
\item we select the maximum modulo: $\max |y^{l\,p}_{r}(t-1)|$ for a recurrent one;  $\max |y^{l\,p}_{r}(t)\cdot 2^{-|l-i|}|$ for a direct\footnote{We use the factor $2^{-|l-i|}$ in order to ensure the priority creation of links with a close layers.} one;
\item if more than one $y^{l\,p}_{r}$ was found, then we will choose any random one of them.
\end{itemize}
If we managed to find some suitable $l, p, r$, then we will go to the final step 5.

\underline{Step 5.} For the chosen $k$ and $y^{l\,p}_{r}$, we assume $c^{i\,j}_{k}(t+1) = (l,p,r)$ and perform an initialization:  $ \omega^{i\,j}_{k}(t+1) = \omega_{\min}$, if $\delta^{i\,j}(t) \leq 0$, and $ \omega^{i\,j}_{k}(t+1) = - \omega_{\min}$, if $\delta^{i\,j}(t) >0$.

\textbf{Algorithm of a redundant link deletion.} On each iteration of a neuron  we make the following steps:\\
\underline{Step 1.} Check if a neuron $N^{i\,j}_{\ldots c}$ has $| \omega_{k}^{i\, j}|$ lower than  $\omega_{\min}$ during $t_{o}$ iterations:
\begin{gather*}\label{nazarov:neuro2:connection_del}
	C^{i\,j}_{del \, k}(t) = 1, \quad \text{if} \,\, \left|  \sum_{t-t_{o}\leq \tau \leq t}  \left(|\omega_{k}^{i\, j}(\tau)|-\omega_{\min}\right) \cdot \sigma(a^{i\,j}(\tau)) \right| < 0.
\end{gather*}

\underline{Step 2.} Delete all connections with $C^{i\,j}_{del \, k}(t) = 1$, assuming $c_{k}^{i\,j}(t+1) = (0,0,0)$. An exception to this rule will be a connection to the external data source $c_{k}^{i\,j}(t) = (0,0,r)$, and also the previously deleted one $c_{k}^{i\,j}(t) = (0,0,0)$, for which our algorithm of a new link creation has found $l,p,r$ on the current iteration $c^{i\,j}_{k}(t+1) = (l,p,r)$.

Compared with previous models, we add  the following parameters:
\begin{itemize}
\item $x_{\max}$ --- maximum absolute value for input data of the neuron;
\item $t_{o}$ --- control time for an old link deletion;
\item $P_{deep1}$ --- probability of creating a deep link bypassing a previous layer;
\item $P_{rec}$ --- probability of a new deep link to be a recurrent one.
\end{itemize}
We can suggest the following general guidelines for selecting these parameters.
The choice of a probability  $P_{rec}$ very much depends on the desired topology of the neural network. Values of $ P_ {rec}> 0.5 $ are selected if the network has a hybrid architecture and prefers creating recurrent connections, while $ P_ {rec} <0.5 $ is chosen if, on the contrary, it is preferable to create deep direct links.
For a $P_{deep1}$ probability, large values should not be chosen, preferably $P_{deep1} \leq 0.2 $.
The control time $t_{o}$ should be chosen small enough $t_{o} \approx 3 \ldots 5$, because otherwise a neuron may fail to reorganize connections in time, which can lead to a paralysis of weight coefficients.
The choice of $x_{\max}$ depends heavily on how input data of neurons was normalized. Most often, we assume $x_{\max}=1$.\\
For the optimal link creation, it is advisable to alternate a supply of training examples in all possible variants of their sequential submission to a neural network.
 In theory, an algorithm  with adaptive connection readjustment can solve the problem with overfitting in deep neural networks. The idea is that it should start its operation almost completely devoid of any connections and with a maximally generalizing  output function. New links are added during the course of training, which leads to a~gradual decrease in a degree of generalization of training examples.


\section{Examples of building a neural network system}

 All the neural networks from our examples will have:
\begin{itemize}
\item external data inputs $\overline{X}(t) = ( X_{1}(t),...,X_{n}(t))$;
\item external data outputs $\overline{Y}(t) = (Y_{1}(t),...,Y_{m}(t))$;
\item reference inputs $\overline{E}(t) = (E_{1}(t),...,E_{m}(t))$;
\item a general training control signal $a(t)$;
\item a general detection of local minimum $\xi(t)$;
\item a general detection of paralysis $p(t)$.
\end{itemize}
The training control signal $a(t)$  will be applied to all the neurons $N^{i\,j}$ from our network as ${a^{i\,j}(t) = a(t)}$. The general detection signals $\xi(t)$ and  $p(t)$ will be constructed with a logical disjunction $\xi(t) = \vee \xi^{i\,j}(t)$ and $p(t) = \vee p^{i\,j}(t)$.
 
 \textbf{\underline{ Example 1}}:Long sort-term memory network with integrated training. Our LSTMIT (LSTM + Integrated Training) network will consist of 9 layers ($j=\overline{1,m}$):
\begin{enumerate}
	\item $N^{1\, j}_{\mathrm{th} \, r}$ --- recurrent input layer. According to our notation this neurons will use equations from model 2 partially replaced by \eqref{nazarov:neuro2:stackmem}--\eqref{nazarov:neuro2:omegarecurrent}.
	\item $N^{2\, j}_{\sigma \, r}$ --- recurrent input gates. They will use equations from model 1 changed by analogy to \eqref{nazarov:neuro2:stackmem}--\eqref{nazarov:neuro2:omegarecurrent}.
	\item $B^{3\, j}_{*}$ --- multiplier blocks (see model 9).
	\item $N^{4\, j}_{\sigma \, r}$ --- recurrent forget gates. They are completely analogous to input gates.
	\item $B^{5\, j}_{* \, r}$ ---recurrent multiplier blocks. Will use equations of model 9 changed by analogy with \eqref{nazarov:neuro2:stackmem}--\eqref{nazarov:neuro2:deltarecurrent}
	\item $B^{6\, j}_{+}$ --- summation blocks (see model 10).
	\item $B^{7\, j}_{\mathrm{th}}$ --- tangent activation blocks (see model 11).
	\item $N^{8\, j}_{\sigma \, r}$ --- recurrent output gates. They are fully analogous to gates of layer 2 and 4.
	\item $B^{9\, j}_{* \, r e}$ --- recurrent output multiplier blocks with  reference inputs $e^{9\, j}(t) = E_{j}(t)$. They will use equations of model 10, changed by analogy with \eqref{nazarov:neuro2:stackmem-e}--\eqref{nazarov:neuro2:stackmem-deltae}.
\end{enumerate}
Since we have already described the operation of all the models considered,    it would be sufficient for a full description to define only static connections $c^{i\,j}$ for all the layers.
\begin{itemize}
	\item Gate and input layers: $c_{k}^{1\, j} = c_{k}^{2\, j} = c_{k}^{4\, j} = c_{k}^{8\, j} = \left\lbrace \begin{aligned}
	(0,0,k), & \quad \text{if } \, k=\overline{1,n},\\
	(9,k,1), & \quad \text{if } \, k=\overline{n+1,n+m}. \end{aligned}  \right.$
	\item Multiplier blocks $B^{3\, j}_{*}$ of layer 3: $c^{3\, j}_{1} = (1,j,1)$ and $c^{3\, j}_{2} = (2,j,1)$.
	\item Multiplier blocks $B^{5\, j}_{* r}$ of  layer 5: $c^{5\, j}_{1} = (4,j,1)$ and $c^{5\, j}_{2} = (6,j,1)$.
	\item Summation blocks $B^{6\, j}_{+}$ of  layer 6: $c^{6\, j}_{1} = (3,j,1)$ and $c^{6\, j}_{2} = (5,j,1)$.
	\item Tangent blocks $B^{7\, j}_{\mathrm{th}}$ of  layer 7: $c^{7\, j}_{1} = (6,j,1)$.
	\item Multipliers $B^{9\, j}_{* r e}$ of the last layer: $c^{9\, j}_{1} = (7,j,1)$ and $c^{9\, j}_{2} = (8,j,1)$.
\end{itemize}

To prepare LSTMIT network for training on $m\leq MaxM$ etalon values, we will have to supply this data values $\overline{X}(t),\ldots , \overline{X}(t+m)$ with corresponding reference values $\overline{E}(t),\ldots , \overline{E}(t+m)$, while holding the control signal low: $a(t) = \ldots = a(t+m) = 0$. To complete one full cycle of training we will need to turn the training signal high  and hold it for an additional $m$ time steps ${a(t+m+1) =}$ ${= \ldots = a(t+2m)= 1}$.

Neurons with adjustable connections could be used for  models, composed of many LSTMIT networks.
However, this should be done only to LSTMIT networks without external data connections and only to their first layers, changing $N^{1\, j}_{\mathrm{th} \, r}$ to $N^{1\, j}_{\mathrm{th} \, r\, c}$ and assuming $P_{deep1} = P_{rec} =0$. At the initial time at least half of the direct links of this layer $N^{1\, j}_{\mathrm{th} \, r\, c}$ should be disabled to suppress the overfitting. In addition to this, we will also have to forbid a deletion of recurrent links, in order to prevent a~disruption of a base LSTM logic.

\textbf{Remark 2}: The main difference of our LSTMIT from classical LSTM is in the introduction of the stack memory  $S_{x_{k}}^{i\, j}$ for $n$ external data inputs of $N^{1\, j}_{\mathrm{th} \, r}$, $N^{2\, j}_{\sigma \, r}$, $N^{4\, j}_{\sigma \, r}$ and $N^{8\, j}_{\sigma \, r}$, as well as the stack memory $S_{e}^{9\, j}$ for external reference inputs of $B^{9\, j}_{* \, r e}$.
It is important to note, that for a model composed of many consequential LSTMIT networks stack memory is not required for inner LSTMIT without external connections.


 \textbf{\underline{ Example 2}}: Radial basis functions with integrated training. This RBFIT network will consist of three layers ($j = \overline{1,m}$):
\begin{enumerate}
	\item[1)] $N^{1\, j}_{\mathrm{Ed}  }$ --- Euclidean distance neurons;
	\item[2)] $B^{2\, j}_{\mathrm{norm}  }$ --- Gaussian activators;
	\item[3)] $N^{3\, 1}_{\mathrm{id}\, e }$ --- single linear neuron with reference input $e(t)$.
\end{enumerate}
All static connections are very simple and straightforward:
\begin{itemize}
	\item for all the input neurons $N^{1\, j}_{\mathrm{Ed}  }$ we have: $c^{1\, j}_{k} = (0,0,k)$, where $k=\overline{1,n}$;
	\item Gaussian activators $B^{2\, j}_{\mathrm{norm}  }$ are linked directly: $c^{2\, j}_{1} = (1,j,1)$, where $j=\overline{1,m}$;
	\item $N^{3\, 1}_{\mathrm{id}\, e }$ is connected to all the second layer: $c^{3\, 1}_{k} = (2,k,1)$, where $k=\overline{1,m}$.
\end{itemize}

It is possible to use $N^{3\, 1}_{\mathrm{id}\, c\, e }$ with adjustable connections instead of $N^{3\, 1}_{\mathrm{id}\, e }$. In this variant we will connect it at the initial time to at least half of the neurons from layer 2, and assume $P_{deep1}   = P_{rec} =0$. As a result, our  network will start with a high degree of input generalization and will gradually decrease it during a training process.


 \textbf{\underline{ Example 3}}: Recurrent radial basis network for chaotic series ($j = \overline{1,m}$):
\begin{enumerate}
	\item[1)] $B^{1\, j}_{+ \, r }$ --- recurrent summation blocks;
	\item[2)] $N^{2\, j}_{\mathrm{Ed}  }$ --- Euclidean distance neurons;
	\item[3)] $B^{3\, j}_{\mathrm{norm}  }$ --- Gaussian activators;
	\item[4)] $N^{4\, 1}_{\mathrm{th} \,r\, e }$ --- hyperbolic tangent neuron with reference input $e(t)$.
\end{enumerate}
This network has only one data input $X_{1}(t)$, and all static connections are very simple:
\begin{itemize}
	\item for the first recurrent layer $c^{1\, j}_{1} = (0,0,1)$ and $c^{1\, j}_{2} = (4,1,1)$ for all $j=\overline{1,m}$;
	\item second layer neurons $N^{2\, j}_{\mathrm{Ed}  }$ are linked directly: $c^{2\, j}_{1} = (1,j,1)$, where $j=\overline{1,m}$;
	\item Gaussian blocks $B^{3\, j}_{\mathrm{norm}  }$ are also linked directly: $c^{3\, j}_{1} = (2,j,1)$, where $j=\overline{1,m}$;
	\item $N^{4\, 1}_{\mathrm{th} \,r\, e }$ is connected to all the third layer: $c^{4\, 1}_{k} = (3,k,1)$, where $k=\overline{1,m}$.
\end{itemize}
The stack memory $S^{1\, j}_{x_{1}}(t)$ for data inputs will be used only for  blocks $B^{1\, j}_{+ \, r }$ and the stack memory for reference inputs $S^{4\, 1}_{e}(t)$ would be used only for a single neuron $N^{4\, 1}_{\mathrm{th} \,r\, e }$.
A training algorithm for the approximation of chaotic sequence  $Y(0),Y(1),\ldots , Y(n)$ will be composed of two steps. At first we will send $X_{1}(t_{0}) = Y(0)$, $X_{1}(t_{0}+\tau) = 0$ to data inputs and
$E_{1}(t_{0}) = Y(1)$, $E_{1}(t_{0}+\tau) = Y(\tau +1)$ to reference inputs, while holding the training signal low: $a(t_{0}+\tau) = 0$ for $\tau = \overline{1,n-1}$. After that we will set the training signal to one and wait additional $n$ cycles: $a(t_{0}+\tau) = 1$ for $\tau = \overline{n,2n}$.

 It is possible to use $N^{4\, 1}_{\mathrm{th} \,r\,c\, e }$ with adjustable links instead of $N^{4\, 1}_{\mathrm{th} \,r\, e }$. In this variant we will connect it at the initial time to only one neuron from layer 2, and assume $P_{deep1} = P_{rec} =0$. As a result parallel approximation branches will be added only when they are necessary, which will gradually increase the probability of successful training.


 \textbf{\underline{ Example 4}}: Convolutional neural networks CONVIT with integrated training. In base variant this network consists of:
\begin{enumerate}
	\item[1)] $N^{1\, j}_{\mathrm{Conv}}$ --- convolutional layer, where $j=\overline{1,n_{1}}$ and all neurons have $m_{1}$ outputs;
	\item[2)]  $B^{2\, j}_{\mathrm{ReLu}}$ --- linear rectification layer, where $j=\overline{1,n_{1} m_{1}}$;
	\item[3)] $B^{3\, j}_{\mathrm{Pool}}$ --- pooling layer, where $j=\overline{1, n_{1} m_{2}}$ and $m_{2} \ll m_{1}$;
	\item[4)] $N^{4\, j}_{\sigma e}$ --- output sigmoid layer, where $j=\overline{1,  n_{1} m_{3}}$ and $m_{3} \ll m_{2}$.
\end{enumerate}
In general case, we assume that the input data is $I_{l\, w\, h}$ in the form of a 3D matrix ($ h = 1 $ for monochrome and $ h = 3 $ for colour). We transform this three-dimensional matrix to a vector form according to a standard algorithm:
\[
	X_{\alpha + l(\beta -1) + lw(\gamma -1)} = i_{\alpha \beta \gamma}, \quad  \, \alpha=\overline{1,l}\quad \beta = \overline{1,w}\quad \gamma = \overline{1,h}.
\]
In addition to this, we will need to introduce the following three auxiliary functions for working with indices ($ x\, \mathrm{rest}\, y $ is the remainder of $x$ divided by $y$; $ [x] $ is the integer part of $x$):
\[
	\theta(x,y) = \left\lbrace \begin{aligned} y, \quad & \text{if }\, y | x,\\ x\, \mathrm{rest}\, y,\quad & \text{if otherwise};  \end{aligned}  \right.
	\]
\[
	\lambda^{+}(x,y) = \left\lbrace \begin{aligned} \left[ \sfrac{x}{y} \right] , \quad & \text{if }\, y | x,\\ \left[ \sfrac{x}{y} \right] +1,\quad & \text{if otherwise};  \end{aligned}  \right.
\]
\[
	\lambda^{-}(x,y) = \left\lbrace \begin{aligned} \left[ \sfrac{x}{y} \right]-1, \quad & \text{if }\, y | x,\\ \left[ \sfrac{x}{y} \right],\quad & \text{if otherwise}.  \end{aligned}  \right.
\]
We assume that  convolutional neurons $N^{1\, j}_{\mathrm{Conv}}$ will read the input data with a sliding window of size $ f \cdot f $ from each slice of a three-dimensional matrix $I_{l\, w\, h}$. Also, we restrict that this window will move along the image with a unit step $ st = 1 $.
After a convolution operation, we will get $m_{1} = l_{1}\cdot w_{1} = (l-f+1)\cdot(w-f+1)$ data at the outputs of each convolutional neurons.
Then these values will pass through the linear rectification blocks and enter the pooling layer.
We choose a pooling window to be of size $ g \cdot g $, which finally yields $m_{1} = g^{2} \, m_{2}=g^{2}\,l_{2}\, w_{2} $ and $l_{2} = l_{1}/g$, $w_{2} = w_{1}/g$.
In this case, the organization of network connections can be carried out as follow:
\begin{itemize}
 \item Every $N^{1\, j}_{\mathrm{Conv}}$ will have $c^{1\, j}_{(\alpha-1)m_{1} + \beta}(t) = (0,0,k)$, where $\alpha=\overline{1,f^{2}h}$, $\beta = \overline{1,m_{1}}$ and
\[
	k = \theta(\alpha,f)\! +\! l\left\lbrace \lambda^{+}(\alpha,f)\! -\! 1\right\rbrace + lw \left\lbrace \lambda^{-}(\alpha,f^{2}) \right\rbrace + l \left\lbrace \theta(\beta, w\! -\! f\! +\! 1)\! - \! 1 \right\rbrace+ \lambda^{-}(\beta, w-f+1).
\]
Such large number of terms is responsible for five types of transitions when reading data with a sliding window $ f \cdot f $.
The term $\theta(\alpha,f)$ is responsible for a movement along a separate column of the sliding window, $l\left\lbrace \lambda^{+}(\alpha,f)-1\right\rbrace$ is included for a transition between this columns, and the term $lw \left\lbrace \lambda^{-}(\alpha,f^{2}) \right\rbrace $  for a transition between separate $h$ colour layers.
By analogy, $l \left\lbrace \theta(\beta, w-f+1) -1 \right\rbrace$ is responsible for a horizontal shift of the sliding window and the final term $\lambda^{-}(\beta, w-f+1)$ for its vertical shift.
\item Second layer  $B^{2\, j}_{\mathrm{ReLu}}$ is linked directly  $c^{2\, j}_{1} = (1,\alpha,\beta)$, where:
\[j = (\alpha -1) m_{1} + \beta \quad \text{and}\quad \alpha = \overline{1,n_{1}}, \,\beta = \overline{1,m_{1}}.
\]
\item For $B^{3\, j}_{\mathrm{Pool}}$ we have $c^{3 j}_{k} = (2,\zeta ,1)$,  $j = (\alpha -1)m_{2} + \beta$ and $\alpha=\overline{1,n_{1}}$, $\beta= \overline{1,m_{2}}$,
\[
	\zeta = (\alpha-1)m_{1} + \theta(k,g) + l_{1}\left\lbrace \lambda^{+}(k,g) -1 \right\rbrace + g\, l_{1}\left\lbrace \theta(\beta,w_{2})-1 \right\rbrace +
	g\, \lambda^{-}(\beta,w_{2}), \, k=\overline{1,g^2}.
\]
The first term  in this expression $(\alpha-1)m_{1}$ is responsible for a transition between $n_{1}$ information channels (from $n_{1}$ neurons of the first layer). The next term $ \theta(k,g)$ denotes the movement along the columns of the sliding window $g \cdot g$. By analogy, $g\, l_{1}\left\lbrace \theta(\beta,w_{2})-1 \right\rbrace$ is responsible for a horizontal shift of the sliding window and $g\, \lambda^{-}(\beta,w_{2})$ is used for its vertical shift.
\item The layer $N^{4\, j}_{\sigma e}$  is connected as: $c^{4 j}_{k} = (3,k,1)$ and $k=\overline{1,n_{1}m_{2}}$, $j=\overline{1,n_{1}m_{3}}$.
\end{itemize}
For a considered basic architecture of a convolutional network, each layer is, in fact, performing some manipulation over three-dimensional data.
In particular, the first layer has $h\cdot l\cdot w$ input values, and outputs $n_{1}\cdot l_{1}\cdot w_{1}$ values to the next layer, which are reduced by pooling layer to $n_{1}\cdot l_{2}\cdot w_{2}$.
The main option of scaling such network is to connect successive layers of   pooling and convolution.
In this case, the convolutional layers will increase the depth of the three-dimensional data $h < n_{1} < n_{2}<  \ldots$, and, at the same time, will reduce the length and width of data $l > l_{1} > \ldots$ and $w> w_{1} > \ldots$, while the pooling layers will only reduce the length and width without altering the depth of data.

 It is possible to use $N^{4\, j}_{\sigma \, c\, e }$ with adjustable connections instead of $N^{4\, j}_{\sigma \, e }$. In this variant we will connect it at the initial time to at least half of the neurons from layer 3, and assume $P_{deep1}   = P_{rec} =0$. As a result, a network will start with a high degree of input generalization and will gradually decrease it during the training process. The comparison with dropout algorithm from [\ref{Srivastava2014}] on MNIST data set is presented in a table   below. The overall time for a training of convit networks was constraint to not exceed the corresponding training time for dropout networks for more then 20\,\%.
 
 \begin{table}[h!]
\caption{Comparison with dropout networks on MNIST data set}
  \begin{center}
    \label{tab:table1}
    \begin{tabular}{l|c|c|c|c} 
\textbf{Method} & \textbf{Unit type} & \textbf{Architecture} & \textbf{Error \%} & \textbf{Time} \\
      \hline
      Dropout NN  & Logistic & 3 layers, 1024 units & 1.35  & $t_{1}$ \\
      Convit NN  & Logistic & 3 layers, 1024 units & 1.38  & $\leq 1.2\, t_{1}$ \\
      \hline
       Dropout NN + max-norm & RELU & 3 layers, 1024 units & 1.06  & $t_{2}$ \\
      Convit NN + max-norm  & RELU & 3 layers, 1024 units & 1.1  & $\leq 1.2\, t_{2}$ \\
      \hline
    \end{tabular}

  \end{center}
\end{table}

Considering the high effectiveness of dropout algorithm we can also add its support in our neuron models. For this we will have to add some variables to store a dropout state:

 \begin{itemize}
\item $\overline{r}^{i\, j}(t) = ({r}^{i\, j}_{1}(t),\ldots,{r}^{i\, j}_{k}(t))$ --- dropout state values for outputs of $N^{i\, j}_{\ldots}(t)$.
\end{itemize}
Updating this variables will be according to a standard rule (probability $p^{i\, j}$ is included in neuron parameters):
\[
	\forall\, i,j,l\quad r^{i\, j}_{l} (t) \sim \mathrm{Bernoulli}(p^{i\, j}).
\]
For example, if we would like to incorporate dropout in model 1, then we will have to change only the formulas for the output ${y}^{i\, j}(t) $ and general correction factors ${\delta}^{i\, j}(t)$:
\[
	{y}^{i\, j}(t) = \left\lbrace \begin{aligned} \varphi^{i\, j} \left( \sum \omega^{i\, j}_{k}(t)\cdot x^{i\, j}_{k}(t) + b^{i\, j}(t) \right)\!, \quad & \text{if }\, a^{i\,j}(t) = 0;\\
	  {r}^{i\, j}(t) \cdot \varphi^{i\, j} \left( \sum \omega^{i\, j}_{k}(t)\cdot x^{i\, j}_{k}(t) + b^{i\, j}(t) \right)\!, \quad & \text{if }\, a^{i\,j}(t) = 1.  \end{aligned}  \right.
\]
 \begin{gather*}\label{nazarov:neuro2:dropoutdelta}
	\delta^{i\, j}(t) =  \left\lbrace  \begin{aligned} (y^{i\, j}(t) - e^{i\, j}(t)), \quad & \text{for  neurons with}\,\, e^{i\,j};\text{ denote them by }  N^{i\, j}_{\sigma e}; \\
	\sum_{ \substack{l,p,k:\\ c^{l\, p}_{k}(t)=(i,j,1)  }} \Delta^{l\, p}_{k}(t)\, {r}^{i\, j}(t), \quad & \text{for  neurons without}\,\, e^{i\,j}; \text{ denote them by }  N^{i\, j}_{\sigma}.  \end{aligned} \right.
 \end{gather*}
The integration of dropout in all the other neuron models will follow a similar scheme with a sole exception of convolutional neurons. For them, we will have to use a dot product for vector  output $\overline{y}^{i\, j}(t)$ and also incorporate dropout coefficients ${r}_{p}^{i\, j}(t)$ in formula \ref{nazarov:neuro2:omega3} as:
  \begin{gather*}\label{nazarov:neuro2:omega3dropout}
	\omega_{k}^{i\, j}(t+1) = \omega_{k}^{i\, j}(t)  -\mu \sum_{m} x^{i\,j}_{k m}   \left\lbrace  \begin{aligned}  (y_{m}^{i\, j}(t) - e_{m}^{i\, j}(t)),    \quad & \text{for} \,\, N^{i\, j}_{\mathrm{Conv}\, e}; \\
	  \! \sum_{ \substack{l_{1},l_{2},p: \\ c^{l_{1}\, l_{2}}_{p}(t)=(i,j,r)  }} \Delta^{l_{1}\, l_{2}}_{p}(t) \, {r}_{m}^{i\, j}(t),  \quad & \text{for} \,\, N^{i\, j}_{\mathrm{Conv}}. \end{aligned} \right.
 \end{gather*}
 

\textbf{\underline{ Example 5}}: Multilayer perceptron PERCIT with integrated training and link adjustment. In a~basic configuration, this neural network consists of $ k $ layers:
\begin{enumerate}
	\item $N^{1\, j}_{\sigma}$ --- input layer, where $j=\overline{1,n_{1}}$ and all neurons have $n$ data inputs.
	\item $N^{2\, j}_{\sigma\, c}$ --- layer with adjustable links, where $j=\overline{1,n_{2}}$ and neurons have $n_{1}$ inputs.
	\item[$k-1$.] $N^{k-1\, j}_{\sigma\, c}$ --- adjustable layer, where $j=\overline{1,n_{k-1}}$ and all neurons have $n_{k-2}$ inputs.
	\item[$k$.] $N^{k\, j}_{\sigma\, c\, e}$ --- output layer, where $j=\overline{1,n_{k}}$ and all neurons have $n_{k-1}$ inputs.
\end{enumerate}
The initial connections will be organized as follow:
\begin{itemize}
	\item all neurons $N^{1\, j}_{\sigma}$ are connected to all the external inputs $c^{1\, j}_{k} = (0,0,k)$, $k=\overline{1,n}$;
	\item for all other layers, we set $75\%$ of all connections to be blank $c^{i\, j}_{k}=(0,0,0)$, and other 25\,\% to be linked to random neurons from the previous layer.
\end{itemize}
For the adjustable neurons, we set the following parameters: the probability of a recurrent connection $P_{rec} = 0$, the probability of creating a deep link bypassing the previous layer $P_{deep_{1}} = 0.1$, the control time for deleting the old links  $t_{o} = 4$ and the maximum absolute value for  input data $x_{\max}=1$.
For deep neural networks, one of the main problems of training them with gradient methods is the vanishing gradient problem.
However, if we allow the creation of deep links with a 10\,\% probability, then we will significantly reduce this effect by passing through the error via several layers.

\textbf{Remark 3}:
 For the optimal creation of new connections, it is advisable to alternate the supply of training examples in all possible variants of their sequential submission. In this case the condition~\ref{nazarov:neuro2:connection_new2} will be used to its full potential.


\section*{Results and Discussion}

An important methodological advantage of our approach is standardization with the development of a universal general formalism for a broad range of neuron models.
First of all, this greatly simplifies an integration of any new models with other activation functions or aggregation of the input data.
Secondly, our approach enables us to construct a hierarchical networks $Nnet_{1},\ldots , Nnet_{k}$, where each $Nnet_{j}$ is controlling the training process of the next network $Nnet_{j+1}$. For example, the first network $Nnet_{1}$ could be trained to spot some basic visual stimuli in video data, which will be used to issue training command for a much bigger second network $Nnet_{2}$. On its part this network $Nnet_{2}$ could be trained with the assistance of the $Nnet_{1}$ to spot a more complex training stimuli, maybe not only in the video data, but in the additional audio data supplied (like verbal command: ``train please'') and learn to associate the corresponding data and issue the training signal for the next one $Nnet_{3}$, and so on.

All of the basic neuron models considered can be easily generalized by switching from standard stochastic gradient descent to a more advanced algorithm. For example, one can integrate stochastic descent with momentum in that models, or stochastic descent with adaptive momentum estimation.


\bigskip


\begin{thebibliography}{9}

\bibitem{Dreyfus}\label{Dreyfus}
{S.E.Dreyfus}, Artificial neural networks, back propagation, and the Kelley-Bryson gradient procedure, {\it Journal of Guidance, Control and Dynamics}, {\bf 13}(1990), 926-928.


\bibitem{Broomhead}\label{Broomhead}
{ D.S.Broomhead, D.Lowe}, Multivariable functional interpolation and adaptive networks, {\it Complex Systems}, {\bf 2}(1988), 321-355.


\bibitem{Lecun}\label{Lecun}
{ Y.Lecun, L.Bottou, Y.Bengio, P.Haffner}, Gradient-based learning applied to document recognition, {\it Proceedings of the IEEE}, {\bf 86}(1998), 2278-2324.

\bibitem{Greff}\label{Greff}
{K.Greff, R.K.Srivastava, J.Koutnik, B.R.Steunebrink, J.Schmidhuber}, LSTM: A search space odyssey, {\it IEEE Trans. Neural Netw. Learn. Syst.}, {\bf 28}(2017), 2222-2232.

\bibitem{Chen}\label{Chen}
{ G.Chen}, A Gentle Tutorial of Recurrent Neural Network with Error Backpropagation, {\it CoRR}, {\bf abs/1610.02583}(2016).


\bibitem{Krizhevsky}\label{Krizhevsky}
{ A.Krizhevsky, I.Sutskever, G.E.Hinton}, Imagenet classification with deep convolutional neural networks,
{\it NIPS}, {\bf 419}(2012), 1106-1114.


\bibitem{Girshick}\label{Girshick}
{ R.Girshick, J.Donahue, T.Darrell, J.Malik}, Rich feature hierarchies for accurate object detection and semantic segmentation, 
{\it CVPR}, {\bf 419}(2014).
 

\bibitem{Park}\label{Park}
{ J.Park, I.W.Sandberg}, Universal Approximation Using Radial-Basis-Function Networks, {\it Neural Computation}, {\bf 3}(1991), 246-257.

\bibitem{Pham}\label{Pham}
{ V.Pham, T.Bluche, C.Kermorvant, J.Louradour}, Dropout improves Recurrent Neural Networks for Handwriting Recognition, {\it arXiv:1312.4569} (2013). 
 
 
\bibitem{Graves}\label{Graves}
{ A.Graves}, Generating sequences with recurrent neural
networks, {\it arXiv:1308.0850}.


\bibitem{Sutskever}\label{Sutskever}
{ I.Sutskever, O.Vinyals, Q.V.Le}, Sequence to Sequence Learning with Neural Networks, {\it arXiv:1409.3215} (2014).


\bibitem{Sak}\label{Sak}
{ H.Sak, A.Senior, F.Beaufays}, Long short-term memory recurrent neural network architectures for large scale acoustic modeling,
{\it In Proc. Interspeech}, (2014).

 \bibitem{Fan}\label{Fan}
{ Y.Fan, Y.Qian, F.Xie, F.K.Soong}, TTS synthesis with bidirectional LSTM based recurrent neural networks, 
{\it In Proc. Interspeech}, (2014).

 
 
\bibitem{Donahue}\label{Donahue}
{ J.Donahue, L.A.Hendricks, S.Guadarrama, et al},  Long-term Recurrent Convolutional Networks for Visual Recognition and Description, {\it arXiv:1411.4389}. 

 
\bibitem{NazarNets1}\label{NazarNets1}
{ M.N.Nazarov},  Artificial neural network with modulation of synaptic coefficients, Vestn. Samar. Gos. Tekhn. Univ., Ser. Fiz.-Mat. Nauki [J. Samara State Tech. Univ., Ser. Phys. Math. Sci.] {\bf 2(31)} (2013), 58-71.

 
\bibitem{Maslennikov}\label{Maslennikov}
{ O.V.Maslennikov, V.I.Nekorkin},  Adaptive dynamical networks, {\it UFN} {\bf 187}(2017)  745-756.  
 
 
\bibitem{Srivastava2014}\label{Srivastava2014}
{ N. Srivastava, G. Hinton , A. Krizhevsky, I. Sutskever ,  R. Salakhutdinov },  Dropout: A simple way to prevent neural networks from overfitting, {\it Journal of Machine Learning Research} {\bf 15}(2014)  1929-1958.  





\end{thebibliography}
\end{document}